\def\year{2019}\relax
\documentclass[letterpaper]{article} %DO NOT CHANGE THIS
\usepackage{aaai19}  %Required
\usepackage{times}  %Required
\usepackage{helvet}  %Required
\usepackage{courier}  %Required
\usepackage{url}  %Required
\usepackage{graphicx}  %Required
\usepackage{booktabs} % for professional tables
\usepackage{tabularx}
\usepackage{arydshln}
\usepackage{wrapfig}
\usepackage{amsmath}
\usepackage{amsfonts}
\usepackage{graphicx}
\usepackage{subcaption}
\usepackage{algorithm}
\usepackage{algorithmic}
\usepackage{latexsym}
\usepackage{multirow}
\usepackage{tablefootnote}
\usepackage{caption}

\begin{document}
\title{Adversarial Bootstrapping for Dialogue Model Training}
% \author{Anonymous Authors}
\author{Oluwatobi O. Olabiyi,
Erik T. Mueller, 
Christopher Larson, 
Tarek Lahlou \\
Capital One Conversation Research, Vienna, VA\\
\{oluwatobi.olabiyi, erik.mueller, christopher.larson2, tarek.lahlou\}@capitalone.com
}
% \author{Oluwatobi O. Olabiyi and 
% Erik T. Mueller\\
% Capital One Conversation Research, Vienna, VA\\
% \{oluwatobi.olabiyi, erik.mueller\}@capitalone.com
% }

\maketitle
\begin{abstract}
Open domain neural dialogue models, despite their successes, are known to produce responses that lack relevance, diversity, and in many cases coherence. These shortcomings stem from the limited ability of common training objectives to directly express these properties as well as their interplay with training datasets and model architectures. Toward addressing these problems, this paper proposes bootstrapping a dialogue response generator with an adversarially trained discriminator. The method involves training a neural generator in both autoregressive and traditional teacher-forcing modes, with the maximum likelihood loss of the auto-regressive outputs weighted by the score from a metric-based discriminator model. The discriminator input is a mixture of ground truth labels, the teacher-forcing outputs of the generator, and distractors sampled from the dataset, thereby allowing for richer feedback on the autoregressive outputs of the generator. To improve the calibration of the discriminator output, we also bootstrap the discriminator with the matching of the intermediate features of the ground truth and the generator's autoregressive output. We explore different sampling and adversarial policy optimization strategies during training in order to understand how to encourage response diversity without sacrificing relevance.  Our experiments shows that adversarial bootstrapping is effective at addressing exposure bias, leading to improvement in response relevance and coherence. The improvement is demonstrated with the state-of-the-art results on the Movie and Ubuntu dialogue datasets with respect to human evaluations and BLUE, ROGUE, and distinct n-gram scores.
% , and human evaluation scores.   

\end{abstract}

\section{Introduction}
\label{introduction}

End-to-end neural dialogue models have demonstrated the ability to generate reasonable responses to human interlocutors. 
% \cite{Sutskever2014,Vinyals2015,Li2016,Serban2016,Xing,Serban2017,Serban2017a,Yu2017,Li2017,Che2017,Zhang2017,Xu2017,Zhang2018b,Li2016c,Nakamura2019,Zhang2018,Olabiyi2018,Olabiyi2018b,Olabiyi2019}. 
However, a significant gap remains between these state-of-the-art dialogue models and human-level discourse. The fundamental problem with neural dialogue modeling is exemplified by their generic responses, such as \textit{I don't know}, \textit{I'm not sure}, or \textit{how are you}, when conditioned on broad ranges of dialogue contexts. In addition to the limited contextual information in single-turn \texttt{Seq2Seq} models \cite{Sutskever2014,Vinyals2015,Li2016}, which has motivated the need for hierarchical recurrent encoder decoder (HRED) multi-turn models \cite{Serban2016,Xing,Serban2017,Serban2017a,Olabiyi2018,Olabiyi2019}, previous work points to three underlying reasons why neural models fail at dialogue response generation.

\textit{i) Exposure Bias:} Similar to language and machine translation models, traditional conversation models are trained with the model input taken from the ground truth rather than a previous output (a method known as \textit{teacher forcing} \cite{Williams1989}). During inference, however, the model uses past outputs,~i.e., is used autoregressively. Interestingly, training with teacher forcing does not present a significant problem in the machine translation setting since the conditional distribution of the target given the source is well constrained. On the other hand, this is problematic in the dialogue setting since the learning task is unconstrained \cite{Lowe2015}. In particular, there are several suitable target responses per dialogue context and vice versa. This discrepancy between training and inference is known as \textit{exposure bias} \cite{Williams1989,Lamb2016} and significantly limits the informativeness of the responses as the decoding error compounds rapidly during inference.  Training methods that incorporate autoregressive sampling into model training have been explored to address this \cite{Li2016c,Li2017,Yu2017,Che2017,Zhang2017,Xu2017,Zhang2018b}.

\begin{figure}[t]
%\vskip 0.2in
\begin{center}
\centerline{\includegraphics[width=0.95\linewidth]{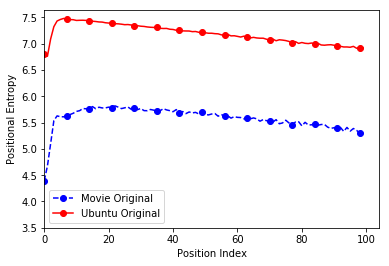}}
\caption{\textbf{Positional Entropy of Movie and Ubuntu datasets -}
Applying a greedy training objective to the datasets can achieve low overall entropy just by overfitting to low entropy regions, resulting in short and generic responses.}
\label{entropy}
\end{center}
\vskip -0.2in
\end{figure}

\textit{ii) Training data:} The inherent problem with dialogue training data, although identified, has not been particularly addressed in the literature \cite{Sharath2017}. Human conversations contain a large number of generic, uninformative responses with little or no semantic information, giving rise to a classic class-imbalance problem. This problem also exists at the word and turn level; human dialogue \cite{Banchs2012,Serban2017} contains non-uniform sequence entropy that is concave with respect to the token position, with the tokens at the beginning and end of a sequence having lower entropy than those in the middle (see Fig. \ref{entropy}). This initial positive energy gradient can create learning barriers for recurrent models, and is a primary contributing factor to their short, generic outputs. 
% In \citeauthor{Shao2017} \shortcite{Shao2017}, the use of glimpse-based decoding is seemingly able to circumvent this problem by breaking this data-induced pattern but at the expense of both training and inference time.

\textit{iii) Training Objective:} Most existing dialogue models are trained using maximum likelihood estimation (MLE) \cite{Sutskever2014,Vinyals2015,Serban2016,Xing} with teacher forcing because autoregressive sampling leads to unstable training. Unfortunately, the use of MLE is incongruent with the redundant nature of dialogue datasets, exacerbates the exposure bias problem in dialogue datasets, and is the primary factor leading to uninteresting and generic responses. Alternative training frameworks that complement MLE with other constraints such as generative adversarial networks, reinforcement learning, and variational auto-encoders that specifically encourage diversity have been explored to overcome the limitations of the MLE objective alone \cite{Li2016,Li2016c,Li2017,Yu2017,Che2017,Zhang2017,Xu2017,Serban2017,Zhang2018b,Olabiyi2018,Olabiyi2019}.  

In this paper, we propose an adversarial bootstrapping framework for training dialogue models. This framework tackles the class imbalance caused by the redundancy in dialogue training data, and addresses the problem of exposure bias in dialogue models. Bootstrapping has been proposed in the literature as a way to handle data with noisy, subjective, and incomplete labels by combining cross-entropy losses from both the ground truth (i.e. teacher forcing) and model outputs (i.e. autoregression) \cite{Reed2015,Grandvalet2005,Grandvalet2006}. Here, we first extend its use to dialogue model training to encourage the generation of high variance response sequences for a given ground truth target \cite{Reed2015}. This should reduce the tendency of dialogue models to reproduce those generic and uninteresting target responses present in the training data. This is achieved by training a discriminator adversarially, and use the feedback from the discriminator to weigh the cross-entropy loss from the model-predicted target. The gradient from the feedback provided by the discriminator encourages the dialogue model to generate a wide range of structured outputs. Second, we bootstrap the discriminator to improve the calibration of its output. We use the similarity between the representations of the generator's autoregressive output and the ground truth from an intermediate layer of the discriminator as an addition target for the discriminator. This further improves the diversity of the generator's output without sacrificing relevance. We apply adversarial bootstrapping to multi-turn dialogue models. Architecture wise, we employ an HRED generator and an HRED discriminator as depicted in Fig. \ref{hred_bootst}, with a shared hierarchical recurrent encoder. In our experiments, the proposed adversarial bootstrapping demonstrates state-of-the-art performances on the Movie and Ubuntu datasets as measured in terms of both automatic (BLUE, ROGUE, and distinct n-gram scores) and human evaluations.

\section{Related Work}
\label{rel_work}

The literature on dialogue modeling even in multi-turn scenario is vast (see \cite{Serban2016,Xing,Serban2017,Serban2017a,Xing,Olabiyi2018,Olabiyi2018b,Olabiyi2019,Li2016c}), and so in this section, we focus on key relevant previous papers. The proposed adversarial bootstrapping is closely related to the use of reinforcement learning for dialogue response generation with an adversarially trained discriminator serving as a reward function \cite{Li2017}. First, we employ a different discriminator training strategy from \citeauthor{Li2017} \shortcite{Li2017}. The negative samples of our discriminator consist of (i) the generator's deterministic teacher forcing output and (ii) distractors sampled from the training set. This makes the discriminator's task more challenging and improves the quality of the feedback to the generator by discouraging the generation of high frequency generic responses. Also, while \citeauthor{Li2017} samples over all the possible outputs of the generator, we take samples from the generator's \textit{top k} outputs or the MAP output with Gaussian noise as additional inputs. This allows our model to explore mostly plausible trajectories during training compared to \citeauthor{Li2017} where the discriminator mostly score the generated samples very low. The \textit{top\_k} sampling strategy also mitigates the gradient variance problem found in the traditional policy optimization employed by \citeauthor{Li2017}. Finally, we bootstrap our discriminator with the similarity between the intermediate representations of the generator's autoregressive output and the ground truth to improve the calibration of the discriminator output. 
% Also, we share similar HRED generator architectures with \cite{Serban2016,Serban2017,Olabiyi2018} and HRED discriminator architectures with \cite{Olabiyi2018}. However, the models in \cite{Serban2016}, \cite{Serban2017}, \cite{Olabiyi2018} are trained using MLE objectives only, combined MLE and variational auto-encoder objectives, and combined MLE and GAN objectives respectively, while we propose a policy-based bootstrapping of MLE objective.

\section{Model}
\label{headings}

Let $\mathbf{x}_i = (x_1, x_2, \cdots, x_i)$ denote the context or conversation history up to turn $i$ and let $x_{i+1}$ denote the associated target response. Provided input-target samples $(\mathbf{x}_i, x_{i+1})$, we aim to learn a generative model $p_{\theta_G}(y_i \mid \mathbf{x}_i)$ which scores representative hypotheses $y_i$ given arbitrary dialogue contexts $\mathbf{x}_i$ such that responses that are indistinguishable from informative and diverse target responses are favored with high scores and otherwise given low scores. Notationally, we write the collection of possible responses at turn $i$ as the set $\mathcal{Y}_i$ containing elements $y_i = (y_i^1, y_i^2, \cdots, y_i^{T_i})$ where $T_i$ is the length of the $i$-th candidate response $y_i$ and $y_i^t$ is the $t$-th word of that response.  

\subsection{Generator Bootstrapping}
\label{gen_boots}
\begin{figure*}[t]
%\vskip 0.2in
\begin{center}
\centerline{\includegraphics[width=0.95\textwidth]{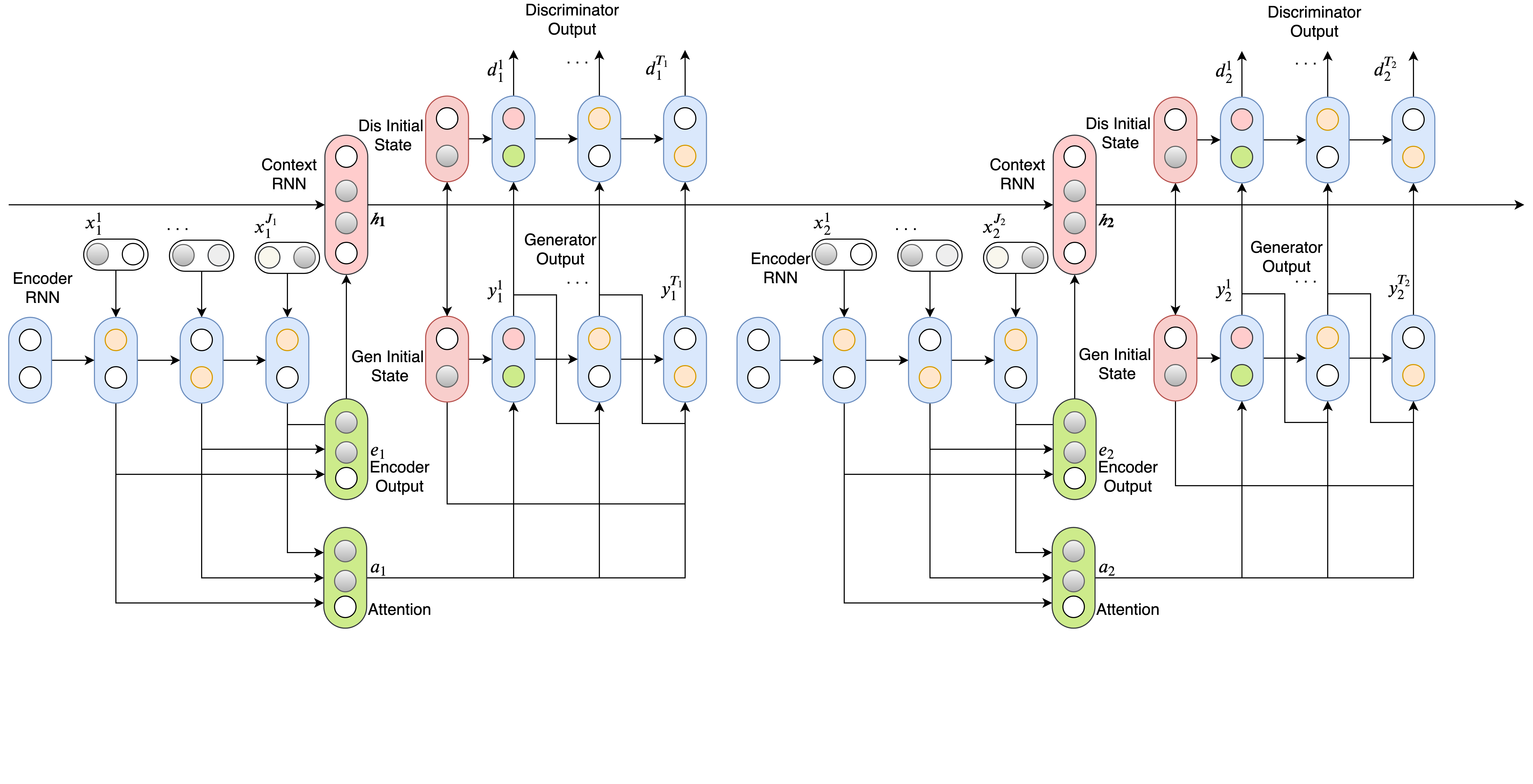}}
\vspace{-35pt}
\caption{\textbf{A multi-turn recurrent architecture with adversarial bootstrapping:} The generator and discriminator share the same encoder (through the context state) and the same word embeddings.  
The generator also uses the word embeddings as the output projection weights. The encoder and the discriminator RNNs are bidirectional while the context and generator RNNs are unidirectional.}
\label{hred_bootst}
\end{center}
% \vskip -0.3in
\vspace{-15pt}
\end{figure*}

To achieve the goal outlined above, we propose an Adversarial Bootstrapping (AB) approach to training multi-turn dialogue models such as the one depicted in Fig.~\ref{hred_bootst}. The adversarial bootstrapping for the generator can be expressed according to the objective
\begin{eqnarray} \label{eq:boot-1}
  \mathcal{L}_{AB}(\theta_G) & = & -\sum_{y_i \in \mathcal{Y}_i} t_G(y_i) \log~p_{\theta_G}(y_i \mid \mathbf{x}_i)
\end{eqnarray}
where $t_G(\cdot)$ is the target variable that controls the generator training. Indeed, hard bootstrapping \cite{Reed2015} is one such special case of \eqref{eq:boot-1} wherein $t_G(y_i) = \beta_{[y_i=x_{i+1}]}$, $t_G(y_i)=1-\beta_{[y_i=\mathop{argmax}\limits_{y_i}p_{\theta_G}(y_i \mid \mathbf{x}_i)]}$, and $0$ otherwise, where $\beta$ is a hyperparameter. Similarly, MLE is another special case in which $t_G(y_i)=1_{[y_i=x_{i+1}]}$, and $0$ otherwise. It is reasonable to assume from these formulations that bootstrapping will outperform MLE since it does not assume all negative outputs are equally wrong.

Interestingly, \citeauthor{Li2017} \shortcite{Li2017} make use of the MLE setting but additionally relies on the sampling stochasticity to obtain non-zero credit assignment information from the discriminator for the generator policy updates. To avoid this inconsistency, we instead modify the generator target to
\begin{eqnarray}
\label{eq:gen_label}
t_G(y_i) & = & \begin{cases}
              \beta   & y_i=x_{i+1} \\
              0       & y_i=\mathop{argmax}\limits_{y_i}p_{\theta_G}(y_i | \mathbf{x}_i) \\
              \alpha Q_{\theta_D}(y_i,\mathbf{x}_i) & otherwise 
          \end{cases}
\end{eqnarray}
where $\alpha$ is a hyperparamter and $Q_{\theta_D}(y_i,\mathbf{x}_i) \in [0, 1]$ is the bootstrapped target obtained from a neural network discriminator $D$ with parameters $\theta_D$. The first two assignments in \eqref{eq:gen_label} are also used in training the discriminator in addition to the human-generated distractors, denoted $x^{-}_{i+1}$, from the dataset. In detail, we make use of the term 
\begin{eqnarray}
\label{eq:dis_label}
t_D(y_i) & = & \begin{cases}
          \beta & y_i=x_{i+1} \\
          0 & y_i=\mathop{argmax}\limits_{y_i}p_{\theta_G}(y_i | \mathbf{x}_i) \\
          0 & y_i=x^{-}_{i+1}
        \end{cases}
\end{eqnarray}
within the context of the objective function. Namely, the discriminator objective is the cross-entropy between the output and the target of the discriminator given by
\begin{multline} \label{eq:boot}
\mathcal{L}_{AB}(\theta_D) = -\sum_{y_i \in \mathcal{Y}_i}[t_D(y_i)\log~Q_{\theta_D}(y_i, \mathbf{x}_i) + \\ (1-t_D(y_i))\log~(1-Q_{\theta_D}(y_i, \mathbf{x}_i))].
\end{multline}
The inclusion of human-generated negative samples encourages the discriminator to assign low scores to high frequency, generic target responses in the dataset, thereby discouraging the generator from reproducing them.
% $D(y_i,\mathbf{x}_i)=\beta_{[y_i=x_{i+1}]}$ and $D(y_i,\mathbf{x}_i)=0_{[y_i=\mathop{argmax}\limits_{y_i}p_{\theta}(y_i | \mathbf{x}_i)]}$.

\subsection{Discriminator Bootstrapping}
\label{dis_boots}
In addition to the generator bootstrapping with the discriminator, we can also bootstrap the discriminator using the similarity measure, $S(.,.) \in [0,1]$, between latent representations of the sampled generator outputs, $y_i \sim p_{\theta_G}(y_i | \mathbf{x}_i)$,  and ground truth encoded by the discriminator. i.e.

\begin{align} \label{eq_dis_boots}
t_D(y_i) = S\big(h_D(y_i), h_D(x_{i+1})\big), y_i \sim p_{\theta_G}(y_i | \mathbf{x}_i)
\end{align}

In our experiments, we chose the cosine similarity metric and the output of the discriminator before the logit projection for $S(.,.)$ and $h_D$ respectively. This helps to better calibrate the discriminator's judgment of the generator's outputs.

\subsection{Sampling Strategy}
\label{sampling} 
To backpropagate the learning signal for the case where $t(y_i)=\alpha Q_{\theta_D}(y_i,\mathbf{x}_i)$, we explore both stochastic and deterministic policy gradient methods. For stochastic policies, we approximate the gradient of $\mathcal{L}_{AB}(\theta_G)$ w.r.t. $\theta_G$ by Monte Carlo samples using the REINFORCE policy gradient method \cite{Li2017,Glynn1990,Williams1992}: 
\begin{multline}
\label{eq:sto_pol}
\nabla_{\theta_G}\mathcal{L}_{AB}(\theta_G) \approx \mathbb{E}_{p_{\theta_G}(y_i | \mathbf{x}_i)}Q_{\theta_D}(y_i,\mathbf{x}_i) ~\cdot \\  \nabla_{\theta_G}\log~p_{\theta_G}(y_i | \mathbf{x}_i).
\end{multline}
For deterministic policies, we approximate the gradient according to \cite{Silver2014,Zhang2018b}
\begin{multline}
\label{eq:det_pol}
\nabla_{\theta_G}\mathcal{L}_{AB}(\theta_G) \approx \mathbb{E}_{p(z_i)}\nabla_{y_{max}}Q_{\theta_D}(y_{max},\mathbf{x}_i) ~\cdot \\  \nabla_{\theta_G}\log~p_{\theta_G}(y_{max} | \mathbf{x}_i, z_i)
\end{multline}
where $y_{max} = \mathop{argmax}\limits_{y_i}p_{\theta_G}(y_i | \mathbf{x}_i, z_i)$ and $z_i \sim \mathcal{N}_i(0,\boldsymbol{I})$ is the source of randomness. We denote the model trained with (\ref{eq:det_pol}) as $aBoots\_gau$.
To reduce the variance of (\ref{eq:sto_pol}), we propose a novel approach of sampling from \textit{top\_k} generator outputs using (i) a categorical distribution based on the output logits ($aBoots\_cat$), similar to the treatment of \citeauthor{Radford2019} \shortcite{Radford2019}, and (ii) a uniform distribution ($aBoots\_uni$); where \textit{top\_k} is a hyperparameter. This is especially useful for dialogue modeling with large vocabulary sizes.

\subsection{Encoder}
\label{encoder}

Referring to the network architecture in Fig.~\ref{hred_bootst}, the generator and discriminator share the same encoder. The encoder uses two RNNs to handle multi-turn representations similar to the approach of \citeauthor{Serban2016} \shortcite{Serban2016}. First, during turn $i$, a bidirectional encoder RNN, $eRNN(\cdot)$, with an initial state of $h_i^0$ maps the conversation context  $x_i$ comprising the sequence of input symbols $(x_i^1, x_i^2,\cdots,x_i^{J_i})$, where $J_i$ is the sequence length, into a sequence of hidden state vectors $\{e_i^j\}_{j=1}^{J_i}$ according to
\begin{eqnarray} \label{eq:encoder_rnn}
e_i^j & = & eRNN(E(x_i^j), e_i^{j-1}), \hspace{2em} j = 1, \cdots, J_i
\end{eqnarray}
where $E(\cdot)$ is the embedding lookup and $E \in \mathbb{R}^{h_{dim} \times V}$ is the embedding matrix with dimension $h_{dim}$ and vocabulary size $V$. 
The vector representation of the input sequence $x_i$ is the $L_2$ pooling over the encoded sequence $\{e_i^j\}_{j=1}^{J_i}$ \cite{Serban2016}. In addition, we use the output sequence as an attention memory to the generator as depicted in Fig.~\ref{hred_bootst}. This is done to improve the relevance of the generated response.

To capture $\mathbf{x}_i$ we use a unidirectional context RNN, $cRNN(\cdot)$, to combine the past dialogue context $\mathbf{h}_{i-1}$ with the $L_2$ pooling of the encoded sequence as
\begin{eqnarray} \label{eq:context_rnn}
  \mathbf{h}_{i} & = & cRNN\left(L_2(\{e_i^j\}_{j=1}^{J_i}), \mathbf{h}_{i-1}\right).
\end{eqnarray} 
% Note that we decided not to allow attention at the turn level, as was done by \citeauthor{Xing} \shortcite{Xing}, since the number of dialogue turns is variable. Also, the use of a single vector representation helps to simplify both training and inference procedures. We however think that converting the turn-level sequential memory to a random access memory \cite{Graves2014,Graves2016} should lead to further system performance improvements without much impact on inference time. We hope to explore this direction in the future.  

\subsection{Generator}
\label{generator}
The generator, denoted $gRNN(\cdot)$, is a unidirectional decoder RNN with an attention mechanism \cite{Bahdanau2015,Luong2015}. Similar to \citeauthor{Serban2016} \shortcite{Serban2016}, the decoder RNN is initialized with the last state of the context RNN. The generator outputs a hidden state representation $g_i^{j}$ for each previous token $\chi^{j-1}$ according to 
\begin{eqnarray} \label{eq:dec_rnn}
g_i^{j} & = &  gRNN\big(E(\chi^{j-1}), g_i^{j-1}, a_i^j, \mathbf{h}_{i}\big)
\end{eqnarray} 
where $a_i^j$ is the attention over the encoded sequence $\{e_i^j\}_{j=1}^{J_i}$. When the generator is run in teacher-forcing mode, as is typically done during training, the previous token from the ground truth is used, i.e., $\chi = x_{i+1}$. During inference (autoregressive mode), the generator's previous decoded output is used, i.e., $\chi = y_i$.

The decoder hidden state, $g_i^{j}$ is mapped to a probability distribution typically through a logistic layer, $\sigma(\cdot)$, yielding,
\begin{align} \label{eq:out_dist}
p_{\theta_G}\big(y_i^j|\chi^{1:j-1},\mathbf{x}_i\big) = softmax(\sigma(g_i^{j})/\tau)
\end{align} 
where $\tau$ is a hyperparameter, $\sigma(g_i^{j})= E \cdot g_i^{j}+b_g$, and $b_g \in \mathbb{R}^{1\times V} $ is the logit bias. 
% Note that the output projection matrix is the same as the embedding matrix, similar to \citeauthor{Vaswani2017} \shortcite{Vaswani2017}.
The generative model can then be derived as:
\begin{align} \label{eq:pf}
p_{\theta_G}(y_i|\mathbf{x}_i) = p_{\theta}(y_i^1|\mathbf{x}_i)\prod_{j=2}^{T_i}p_{\theta_G}\big(y_i^j | \chi^{1:j-1}, \mathbf{x}_i\big)
\end{align}
\subsection{Discriminator}
\label{discriminator}
The discriminator $Q_{\theta_D}(y_i,\mathbf{x}_i)$ is a binary classifier that takes as input a response sequence $y_i$ and a dialogue context $\mathbf{x}_i$ and is trained with output labels provided in (\ref{eq:dis_label}) and (\ref{eq_dis_boots}). 
The discriminator, as shown in Fig. \ref{hred_bootst} is an RNN, $dRNN(\cdot)$, that shares the hierarchical encoder and the word embeddings with the generator, with the initial state being the final state of the context RNN. The last layer of the discriminator RNN is fed to a logistic layer and a sigmoid function to produce the normalized $Q$ (action-value function) value for a pair of dialogue context (state) and response (action).      
% This definition is more general than used by \citeauthor{Li2017} \shortcite{Li2017}, as the target is not just a binary classification of the dialogue context and response pair as either human-generated ($Q_+$) or machine-generated ($Q\_$). That is, the target also includes the distractor classification, and the output of a similarity function.  This formulation allows us to view the discriminator as a metric learner \cite{Kulis2013} as opposed to the generator, which is a maximum likelihood learner. Therefore, adversarial bootstrapping can be viewed as joint metric and maximum likelihood learning.    

We explore two options of estimating the $Q$ value, i.e., at the word or utterance level. At the utterance level, we use (\ref{eq:boot}) in conjunction with a unidirectional discriminator RNN. The $Q$ value is calculated using the last output of $dRNN(\cdot)$, i.e,   
\begin{align} \label{eq:dis_pf_utt}
Q_{\theta_D}(y_i,\mathbf{x}_i) = sigmoid(\sigma(d_i^{T_i}))
\end{align}
where $\sigma(d_i^{T_i}) = W_d \cdot hg_i^{j}+b_d$, $W_d \in \mathbb{R}^{h_{dim} \times V}$, and $b_g \in \mathbb{R}^{1\times V} $ are the logit projection and bias respectively.
At the word level, the discriminator RNN (we use a bidirectional RNN in our implementation) produces a word-level evaluation. The normalized $Q$ value and the adversarial bootstrapping objective function are then respectively given by
\begin{align} \label{eq:dis_pf_token}
d_i^j = Q_{\theta_D}(y_i^j,\mathbf{x}_i|y_i) = sigmoid(\sigma(d_i^j))
\end{align}
\begin{align} \label{eq:boot_token}
\mathcal{L}_{AB}(\theta_G) = -\sum_{y_i \in \mathcal{Y}_i}\sum_{j=1}^{T_i}t_G(y_i^j)log~p_{\theta_G}(y_i^j | \mathbf{x}_i)
\end{align}
where 
\begin{align}
\label{eq:gen_label_token}
t_G(y_i^j) = \begin{cases}
          \beta & y_i^j=x_{i+1}^j \\
          0 & y_i^j=\mathop{argmax}\limits_{y_i^j}p_{\theta_G}(y_i^j | \chi^{1:j-1}, \mathbf{x}_i) \\
          (1-\beta)d_i^j & otherwise 
        \end{cases}
\end{align}

% \addtocounter{footnote}{-1}
% \footnotetext{LM, similar to GPT-2 \cite{Radford2019}, uses only the decoder portion of the original Transformer architecture \cite{Vaswani2017}.}
% \addtocounter{footnote}{1}
% \footnotetext{Seq2Seq uses both the encoder and the decoder portions of the original Transformer architecture \cite{Vaswani2017}.}

\section{Training}
\label{training}
We train both the generator and discriminator simultaneously with two samples for the generator and three for the discriminator. In all our experiments, we use the generator's teacher forcing outputs to train the discriminator (i.e., $argmax$ cases of (\ref{eq:gen_label}) and (\ref{eq:dis_label})). The encoder parameters are included with the generator, i.e., we did not update the encoder during discriminator updates. Each RNN is a 3-layer GRU cell, with a hidden state size ($h_{dim}$) of 512. The word embedding size is the same $h_{dim}$, and the vocabulary size $V$ is $50000$. Other hyperparameters include $\beta=1$, $\alpha=1$, $\tau=1$, $top\_k=10$ for $aBoots\_uni$ and $aBoots\_cat$, and $top\_k=10$ for $aBoots\_gau$. Although we used a single \textit{top\_k} value during training, we avoided training with multiple \textit{top\_k} values by searching for the optimum \textit{top\_k} (between 1 and 20) on the validation set using the BLEU score. We used the obtained optimum values for inference.
Other training parameters are as follows: the initial learning rate is $0.5$ with decay rate factor of $0.99$, applied when the generator loss has increased over two iterations. We use a batch size of 64 and clip gradients around $5.0$. All parameters are initialized with Xavier uniform random initialization \cite{Glorot2010}. Due to the large vocabulary size, we use sampled softmax loss \cite{Jean2015} for the generator to limit the GPU memory requirements and expedite the training process. However, we use full softmax for evaluation. The model is trained end-to-end using the stochastic gradient descent algorithm. 

\begin{table*}[ht]
\caption{Automatic evaluation of generator performance}
\label{tb:auto}
%\vskip 0.15in
% \vspace{5pt}
\begin{center}
\begin{small}
% \begin{sc}
\vspace{-10pt}
\setlength\tabcolsep{4.0pt}
\begin{tabular}{l|cc|cc||cc|cc}
\toprule
\multirow{3}{*}{\textbf{Model}}  &   \multicolumn{4}{c||}{\textbf{Movie}} & \multicolumn{4}{c}{\textbf{Ubuntu}} \\
& \multicolumn{2}{c|}{Relevance} & \multicolumn{2}{c||}{Diversity} & \multicolumn{2}{c|}{Relevance} & \multicolumn{2}{c}{Diversity} \\    
 & BLEU & ROUGE & DIST-1/2 & NASL & BLEU & ROUGE & DIST-1/2 & NASL \\
\midrule
HRED      & 0.0474   & 0.0384 & 0.0026/0.0056 & 0.535 & 0.0177  & 0.0483  & 0.0203/0.0466 & 0.892 \\
VHRED     & 0.0606   & 0.1181 & 0.0048/0.0163 & 0.831 & 0.0171  & 0.0855  & 0.0297/0.0890 & 0.873 \\
hredGAN\_u & 0.0493   & 0.2416 & 0.0167/0.1306 & 0.884 & 0.0137  & 0.0716  & 0.0260/0.0847 & 1.379 \\
hredGAN\_w & 0.0613 & 0.3244 & 0.0179/0.1720 & 1.540 & 0.0216  & 0.1168  & 0.0516/0.1821 & 1.098 \\
DAIM & 0.0155 & 0.0077 & 0.0005/0.0006 & 0.721 & 0.0015 & 0.0131 & 0.0013/0.0048 & \textbf{1.626} \\
% \midrule
% Transformer LM\protect\footnotemark (117M)    & 0.0518   & 0.0630 & 0.0176/0.0540 & 1.101 & 0.0171  & 0.0855  & 0.0297/0.0890 & 0.873 \\
Transformer & 0.0360   & 0.0760 & 0.0107/0.0243 & \textbf{1.602} & 0.0030  & 0.0384  & 0.0465/0.0949 & 0.566 \\
% Transformer LM (345M)     & 0.0042   & 0.0564 & 0.0300/0.0683 & 0.320 & 0.0171  & 0.0855  & 0.0297/0.0890 & 0.873 \\
% Transformer Seq2Seq (345M) & 0.0057 & 0.0970 & 0.1568/0.3785 & 0.331 & 0.0015  & 0.0345  & 0.1555/0.3990 & 0.470 \\
\midrule
aBoots\_u\_gau & 0.0642 & 0.3326 & 0.0526/0.2475 & 0.764 & 0.0115   & 0.2064 & 0.1151/0.4188 & 0.819 \\
aBoots\_w\_gau & 0.0749 & 0.3755 & 0.0621/0.3051 & 0.874 & 0.0107   & 0.1712 & \textbf{0.1695/0.7661} & 1.235 \\
aBoots\_u\_uni & 0.0910  & 0.4015 & 0.0660/0.3677 & 0.975 & 0.0156   & 0.1851 & 0.0989/0.4181 & 0.970 \\
aBoots\_w\_uni & 0.0902  & \textbf{0.4048} & \textbf{0.0672/0.3653} & 0.972 & 0.0143 & 0.1984 & 0.1214/0.5443 & 1.176 \\
aBoots\_u\_cat & 0.0880  & 0.4063 & 0.0624/0.3417 & 0.918 & 0.0210   & 0.1491 & 0.0523/0.1795 & 1.040 \\
\textbf{aBoots\_w\_cat} & \textbf{0.0940}   & 0.3973 & 0.0613/0.3476 & 1.016 & \textbf{0.0233} & \textbf{0.2292} & 0.1288/0.5190 & 1.208 \\

\bottomrule
\end{tabular}
\end{small}
\end{center}
% \vspace{-20pt}
\vspace{-10pt}
\end{table*}

\begin{table*}[ht]
\caption{Human evaluation of generator performance based on response informativeness}
\label{tb:human}
%\vskip 0.15in
% \vspace{5pt}
\begin{center}
\begin{small}
% \begin{sc}
\vspace{-10pt}
\setlength\tabcolsep{12.0pt}
\begin{tabular}{lcc}
\toprule
\textbf{Model Pair}  &   \textbf{Movie} & \textbf{Ubuntu} \\
% \multirow{3}{*}{\textbf{Models}}  &   \multicolumn{2}{c||}{\textbf{Movie}} & \multicolumn{2}{c}{\textbf{Ubuntu}} \\
 % & \multicolumn{2}{c}{Informativeness} & \multicolumn{2}{c}{Informativeness} \\    
% & aBoots\_w\_cat & HRED & aBoots\_w\_cat & VHRED & aBoots\_w\_cat & hredGAN\_w & aBoots\_w\_cat & DAIM \\
\midrule
aBoots\_w\_cat -- DAIM & \textbf{0.957} -- \textbf{0.043} & \textbf{0.960} -- \textbf{0.040} \\
aBoots\_w\_cat -- HRED      & \textbf{0.645} -- \textbf{0.355} & \textbf{0.770} -- \textbf{0.230} \\
aBoots\_w\_cat -- VHRED      & \textbf{0.610}   -- \textbf{0.390} & \textbf{0.746} -- \textbf{0.254} \\
aBoots\_w\_cat -- hredGAN\_w    & 0.550   -- 0.450 & 0.556 -- 0.444 \\

\bottomrule
\end{tabular}
\end{small}
\end{center}
% \vspace{-20pt}
\vspace{-10pt}
\end{table*}

\begin{table*}[h]
\caption{Automatic evaluation of $aBoots$ models with the generator bootstrapping only}
\label{tb:auto_part}
%\vskip 0.15in
% \vspace{5pt}
\begin{center}
\begin{small}
% \begin{sc}
\vspace{-10pt}
\setlength\tabcolsep{4.0pt}
\begin{tabular}{l|cc|cc||cc|cc}
\toprule
\multirow{3}{*}{\textbf{Model}}  &   \multicolumn{4}{c||}{\textbf{Movie}} & \multicolumn{4}{c}{\textbf{Ubuntu}} \\
& \multicolumn{2}{c|}{Relevance} & \multicolumn{2}{c||}{Diversity} & \multicolumn{2}{c|}{Relevance} & \multicolumn{2}{c}{Diversity} \\    
 & BLEU & ROUGE & DIST-1/2 & NASL & BLEU & ROUGE & DIST-1/2 & NASL \\
\midrule
aBoots\_g\_u\_gau & 0.0638 & 0.3193 & 0.0498/0.2286 & 0.778 & 0.0150   & 0.1298 & 0.0480/0.1985 & 0.960 \\
aBoots\_g\_w\_gau & 0.0729 & 0.3678 & 0.0562/0.3049 & 1.060 & 0.0123   & 0.1370 & 0.0646/0.1820 & 0.841 \\
aBoots\_g\_u\_uni & 0.0801  & 0.3972 & 0.0655/0.3414 & 0.869 & 0.0124   & 0.1424 & 0.0636/0.1853 & 0.870 \\
aBoots\_g\_w\_uni & 0.0860  & \textbf{0.4046} & \textbf{0.0671/0.3514} & 0.838 & 0.0170 & 0.2049 & 0.1074/0.4646 & 1.349 \\
aBoots\_g\_u\_cat & 0.0836  & 0.3887 & 0.0597/0.3276 & 0.917 & 0.0131   & 0.1214 & 0.0597/0.3276 & 1.060 \\
aBoots\_g\_w\_cat & 0.0928   & \textbf{0.4029} & \textbf{0.0613/0.3358} & 0.976 & 0.0202 & \textbf{0.2343} & \textbf{0.1254/0.4805} & 0.873 \\
\midrule

\bottomrule
\end{tabular}
\end{small}
\end{center}
% \vspace{-20pt}
\vspace{-10pt}
\end{table*}

\section{Experiments}
\label{experiments}

\subsection{Setup}
We evaluated the proposed adversarial bootstrapping (\textit{aBoots}) with both generator and discriminator bootstrapping, on the \textbf{Movie Triples} and \textbf{Ubuntu Dialogue} corpora randomly split into training, validation, and test sets, using 90\%, 5\%, and 5\% proportions. We performed minimal preprocessing of the datasets by replacing all words except the top 50,000 most frequent words by an \textit{UNK} symbol.
The Movie dataset \cite{Serban2016} spans a wide range of topics with few spelling mistakes and contains about 240,000 dialogue triples, which makes it suitable for studying the relevance vs. diversity tradeoff in multi-turn conversations. The Ubuntu dataset, extracted from the Ubuntu 
Relay Chat Channel \cite{Serban2017}, contains about 1.85 million conversations with an average of 5 utterances per conversation. This dataset is ideal for training dialogue models that can provide expert knowledge/recommendation in domain-specific conversations.

We explore different variants of \textit{aBoots} along the choice of discrimination (either word(\_w) or utterance(\_u) level) and sampling strategy (either uniform(\_uni), categorical(\_cat) or with Gaussian noise (\_gau)). We compare their performance with existing state-of-the-art dialogue models including (V)HRED\footnote{implementation obtained from \url{https://github.com/julianser/hed-dlg-truncated}} \cite{Serban2016,Serban2017}, and DAIM\footnote{implementation obtained from \url{https://github.com/dreasysnail/converse_GAN}} \cite{Zhang2018b}. For completeness, we also include results from a transformer-based Seq2Seq model \cite{Vaswani2017}.  

We compare the performance of the models based on the informativeness (a combination of relevance and diversity metrics) of the generated responses. For relevance, we adopted BLEU-2 \cite{Papineni2002} and ROUGE-2 \cite{Lin2014} scores. For diversity, we adopted distinct unigram (DIST-1) and bi-gram (DIST-2) \cite{Li2016} as well as and normalized average sequence length (NASL) scores \cite{Olabiyi2018}.

For human evaluation, we follow a similar setup as \citeauthor{Li2016} \shortcite{Li2016}, employing crowd sourced judges to evaluate a random selection of 200 samples.
We present both the multi-turn context and the generated responses from the models to 3 judges and ask them to rank the response quality in terms of informativeness. Ties are not allowed. The informativeness measure captures the temporal appropriateness, i.e, the degree to which the generated response is temporally and semantically appropriate for the dialogue context as well as other factors such as length of the response, and repetitions. For analysis, we pair the models and compute the average number of times each model is ranked higher than the other. 
% We also performed statistical significant tests (p<0.001) on each model pair.

\section{Results and Discussion}
\label{res_dis}

\subsection{Quantitative Evaluation}
\label{eval_quant}
The quantitative measures reported in Table \ref{tb:auto} show that adversarial bootstrapping $aBoots$ gives the best overall relevance and diversity performance in comparison to (V)HRED, hredGAN, DAIM and Transformer, on both the Movie and Ubuntu datasets. We believe that the combination of improved discriminator training and the policy-based objective is responsible for the observed performance improvement. On the other hand, multi-turn models (V)HRED and hredGAN suffer performance loss due to exposure bias, since autoregressive sampling is not included in their training. Although DAIM uses autoregressive sampling, its poor performance shows the limitation of the single-turn architecture and GAN objective compared to the multi-turn architecture and policy-based objective in $aBoots$. The transformer Seq2Seq model, which performs better than RNNs on the machine translation task, also suffers from exposure bias, and overfits very quickly to the low entropy regions in the data, which leads to a poor inference performance.  Also, the results from $aBoots$ models indicate that word-level discrimination performs better than utterance-level discrimination, consistent with the results reported by \citeauthor{Olabiyi2018} \shortcite{Olabiyi2018} for the hredGAN model. While it is difficult to identify why some models generate very long responses, we observe that models with Gaussian noise inputs (e.g., hredGAN and $aBoots\_gau$) may be using the latent Gaussian distribution to better encode response length information; indeed, this is an area of ongoing work. Within the variants of $aBoots$, we observe that models trained with a stochastic policy, $aBoots\_uni$ and $aBoots\_cat$, outperform those trained with a deterministic policy, $aBoots\_gau$. Notably, we find that for the stochastic policy, there is a tradeoff in relevance and diversity between \textit{top\_k} categorical and uniform sampling. The categorical sampling tends to perform better with relevance but worse with diversity. We believe that this is because \textit{top\_k} categorical sampling causes the generator to exploit high likelihood (i.e., more likely to be encountered during inference) than uniform sampling of the top candidates, while still allowing the policy to explore. This however comes with some loss of diversity, although not significant. Overall, the automatic evaluation indicates that adversarial bootstrapping trained with stochastic policy using \textit{top\_k} categorical sampling strategy gives the best performance.

% Within $aBoots$, the variants using word-level discrimination performs better than those with utterance-level discrimination, while the ones trained with stochastic policy performs better than with deterministic policy. We note that without the novel proposed \textit{top\_k} sampling strategy, the stochastic policy performs similarly as the deterministic policy. The \textit{top\_k} sampling ensures that generator explores generating responses that are more likely to be encountered during inference. Also, there seems to be a trade-off of relevance and diversity between categorical sampling and uniform sampling.              

\subsection{Qualitative Evaluation}
\label{eval_quanl}
As part of our evaluation we also consider scores from human judges. Specifically, we have each evaluator compare responses from five models: $aBoots\_w\_cat$, hredGAN\_w, (V)HRED, and DAIM. The pairwise human preferences are reported in Table \ref{tb:human}. These data indicate a significant preference for responses generated by
$aBoots\_w\_cat$ as compared to both (V)HRED and DAIM. We observe that $aBoots\_w\_cat$ is preferred over hredGAN\_w on average, although not by a significant margin. We note that this score was computed from only 200 evaluation samples, which is likely too small to demonstrate a strong preference for $aBoots\_w\_cat$. It is also worthy noting that the hredGAN\_w model represents a strong baseline, based on previous human evaluations \cite{Olabiyi2018}, against which to compare our adversarially trained models. It is interesting to note that although automatic evaluation scores hredGAN\_w much lower than $aBoots\_w\_cat$ on relevance, the long response length from hredGAN\_w, which indicates strong diversity, has a considerable impact on how human evaluators judge the informativeness of responses generated by hredGAN\_w. 
Table \ref{tb:samples} shows example responses from the models.

\iftrue
\begin{table}[t]
\caption{ Sample responses of DAIM, (V)HRED, hredGAN, $aBoots\_w\_cat$.}
\label{tb:samples}
%\vskip 0.15in
\vspace{-10pt}
\begin{center}
\begin{tiny}
%\begin{sc}
%\begin{tabular}{lllr}
\setlength\tabcolsep{6.0pt} % default value: 6pt
% \parbox{.45\textwidth}{
\begin{tabularx}{\linewidth}{ l X }
\toprule
%Model & Response \\
\textbf{Model} & \textbf{Response} \\
\textbf{Movie} \\
\toprule
\textbf{Context\_0}& we ' re grateful , $<$person$>$ . it ' s not that . those craftsman tools . they ' re a dream come true .\\
\textbf{Context\_1}& your dad and i were raised with a certain work ethic , that ' s all .\\
% \textbf{DAIM}& you . be would is made and it \_UNK ?\\
\textbf{HRED}& $<$person$>$ .\\
\textbf{VHRED}& $<$person$>$ , mr . $<$person$>$ . $<$person$>$ , i ' m not going to tell you , $<$person$>$ . i ' m sorry . i ' m sorry , $<$person$>$ . i ' m sorry \\
\textbf{hredGAN\_w}& $<$person$>$ , i know you ' re see . $<$ continued\_utterance $>$ $<$person$>$ , please .\\
\textbf{$aBoots\_w\_cat$}& $<$person$>$ , i ' m a little worried about my future . i ' m not here to marry you . \\
\midrule %\hdashline
\textbf{Context\_0}& where would they be going ?\\
\textbf{Context\_1}& we ' ve got two choices . one of them is wrong .\\
% \textbf{DAIM}& you . be would is made and it \_UNK ?\\
\textbf{HRED}& i ' m sorry , $<$person$>$ .\\
\textbf{VHRED}& i know . \\
\textbf{hredGAN\_w}& $<$person$>$ , i know , it ' s a real hair . with you a little little little , little little little little somethin\\
\textbf{$aBoots\_w\_cat$}& i don ' t care where they are , i don ' t know anyone will be outside .\\
% \toprule
% \textbf{Context\_0}& $<$person$>$ i ' ve raised my rate . i might even cut my sessions to forty minutes .\\
% \textbf{HRED}& i ' m sorry , $<$person$>$ .\\
% \textbf{VHRED}& $<$person$>$, $<$person$>$, $<$person$>$, $<$person$>$ . \\
% \textbf{hredGAN\_w}& $<$person$>$ , i ' m sorry , but i ' m not sure if you ' re really a very good guy . \\
% \textbf{hredGAN\_u}&  i ' ll never see them again , $<$person$>$ . \\
% \midrule
% \textbf{Context\_1}& $<$person$>$ , i know you don ' t like flattery but if anybody i know is going to win a nobel prize , it ' s you . you got ta be thinking about your next book so i know you ' ll find this case particularly interesting . \\
% \textbf{HRED}&  i ' m sorry , $<p$erson$>$ . \\
% \textbf{VHRED}:& it ' s not a lot of $<$unk$>$ . i ' m not going to tell you , $<$person$>$ . i ' m not going to help you . \\
% \textbf{hredGAN\_w}& i ' m a little nervous , but i don ' t care if there ' s something more important than $<$person$>$ . \\
% \textbf{hredGAN\_u}& $<$person$>$ , i don ' t know what you ' m talking about .\\
\toprule
\textbf{Ubuntu} \\
\toprule
\textbf{Context\_0}& DJones: update manager won't detect 12.04.1 as a new version if you are already running 12.04, because 12.04.1 = 12.04 + lots of package updates\\
\textbf{Context\_1}& DJones I am running 10.04.\\
\textbf{Context\_2}& what does 'gksu update-manager \-c' report?\\
% \textbf{DAIM}& what : : : .\\
\textbf{HRED}&  I don't know what the problem is \\
\textbf{VHRED}&  you mean ? \\
\textbf{hredGAN\_w}& yes , I get the new one , I just want to get the new one \\
\textbf{$aBoots\_w\_cat$}& it says that command not found \\
\midrule
\textbf{Context\_0}& libgnome2-dev is the package that has that pkgconfig file\\
\textbf{Context\_1}& After I installed libgnome2-dev I got a different output. http://pastebin.com/zpDeGs81 sorry it is so long.\\
% \textbf{Context\_2}& your user doesn't have permission to write to that location, you in general need to: sudo make install; (though I generally prefer to adjust --prefix to install it locally or build a package)\\
\textbf{Context\_2}& I thought that was what was going on thanks. :) I ran the command ./configure \&\& sudo make \&\& sudo make install would that have worked.\\
% \textbf{DAIM}& : : : : ) : ) ) ) ) ) )\\
\textbf{HRED}&   I don't know what the problem is \\
\textbf{VHRED}&  what do you mean ? \_\_eou\_\_ what are you trying ? \\
\textbf{hredGAN\_w }& I have no idea , I can just copy the files and copy files to the other one \\
\textbf{$aBoots\_w\_cat$}& yes I did . I just did sudo apt-get update and it worked\\
\bottomrule
%\end{tabular}
\end{tabularx}
% }
%\end{sc}
\end{tiny}
\end{center}
%\vskip -0.1in
\vspace{-10pt}
% \end{minipage}
\end{table}
% \end{tabular}
\fi
\subsection{Ablation Studies}
In this section, we examine the effect of partial bootstrapping on the model performance. Here, the target in (\ref{eq_dis_boots}) is excluded from the discriminator. The automatic evaluation results on all the variants of $aBoots$ are reported in Table \ref{tb:auto_part}. The table shows that the generator models bootstrapped by a discriminator that is not bootstrapped generally performs worse than ones with a bootstrapped discriminator. This improvement is particularly more evident in the best performing variant, $aBoots\_w\_cat$. We attribute this performance improvement to the better calibration of discriminator obtained from the bootstrapping of the discriminator output with the similarity measure between the generator's autoregressive output and the ground truth during training.     

\section{Conclusion}
\label{concl}
We have proposed a novel training technique, adversarial bootstrapping, which is useful for dialogue modeling. The method addresses the issues of data-induced redundancy and exposure bias in dialogue models trained with maximum likelihood. This is achieved by bootstrapping the teacher-forcing MLE objective with feedback on autoregressive outputs from an adversarially trained discriminator. This feedback discourages the generator from producing bland and generic responses that are characteristic of MLE training. Experimental results indicate that a doubly bootstrapped system produces better performance than a system where only the generator is bootstrapped. Also, the model variant characterized by choosing \textit{top\_k} categorical sampling, stochastic policy optimization, and word-level discrimination gives the best performance.
The results demonstrate that the proposed method leads to models generating more relevant and diverse responses in comparison to existing methods.

% \clearpage
% \break
\bibliography{aaai_2020_adv_bootstrapping}
\bibliographystyle{aaai}

\end{document}